\documentclass[10pt,twocolumn,letterpaper]{article}

\usepackage{wacv}
\usepackage{times}
\usepackage{epsfig}
\usepackage{graphicx}
\usepackage{amsmath}
\usepackage{amssymb}
\usepackage{booktabs}
\usepackage{booktabs}
\usepackage{multirow}
\usepackage{amssymb}
\usepackage{cite}
\usepackage{algorithm}
\usepackage{algorithmic}
\usepackage[table]{xcolor}
\usepackage{xspace}

\usepackage[pagebackref,breaklinks,colorlinks]{hyperref}

\newcommand{\inmotion}{\mathbf{x}_{1:T}}
\newcommand{\motion}{x}
\newcommand{\speed}{v}
\newcommand{\futuremotion}{\mathbf{x}_{T+1:T+N}}
\newcommand{\predmotion}{\mathbf{x'}_{T+1:T+N}}
\newcommand{\futurespeed}{\mathbf{v}_{T+1:T+N}}
\newcommand{\predspeed}{\mathbf{v'}_{T+1:T+N}}
\renewcommand{\paragraph}[1]{\noindent\textbf{#1}\quad}

\usepackage{dsfont}
\usepackage{etoolbox}

\newif\ifshowedits

\newcommand{\addeditor}[3]{%
  \definecolor{#1color}{rgb}{#3}
  \expandafter\newcommand\csname #1\endcsname[1]
{\ifshowedits{\color{#1color} ##1}\else{##1}\fi}%
  \expandafter\newcommand\csname #1rmk\endcsname[1]
{\ifshowedits{\color{#1color}{\bf [#2: ##1]}}\fi}%
  \expandafter\newcommand\csname #1rpl\endcsname[2]{%
  \ifshowedits
    {\color{#1color} ##1 \sout{##2}}
  \else
    {##1}
  \fi
  }%
}

\newcommand{\createtextvar}[1]{
  \expandafter\newcommand\csname #1\endcsname{%
  {\text{#1}}
}%
}
\newcommand{\textvars}[1]{\forcsvlist{\createtextvar}{#1}}

\newcommand{\mycomment}[1]{}

\DeclareFontFamily{U}{mathx}{\hyphenchar\font45}
\DeclareFontShape{U}{mathx}{m}{n}{
      <5> <6> <7> <8> <9> <10>
      <10.95> <12> <14.4> <17.28> <20.74> <24.88>
      mathx10
      }{}
\DeclareSymbolFont{mathx}{U}{mathx}{m}{n}
\DeclareFontSubstitution{U}{mathx}{m}{n}
\DeclareMathAccent{\widebar}{0}{mathx}{"73}

\showeditstrue

\textvars{DCT,IDCT,GRU,BN,GCN}

\newcommand{\method}{\textsc{siMLPe}\xspace}

\def\wacvPaperID{381}

\wacvalgorithmstrack

\wacvfinalcopy
\usepackage[accsupp]{axessibility}  

\begin{document}

\title{Back to MLP: A Simple Baseline for Human Motion Prediction}

\author{Wen Guo$^{1}$\footnotemark[1], Yuming Du$^{3}$\footnotemark[1], Xi Shen$^{4}$\footnotemark[2], Vincent Lepetit$^{3}$, 
Xavier Alameda-Pineda$^{1}$, Francesc Moreno-Noguer$^{2}$ \\
$^1$ Inria, Univ. Grenoble Alpes, CNRS, Grenoble INP, LJK, 38000 Grenoble, France \\
$^2$ Institut de Robòtica i Informàtica Industrial, CSIC-UPC, Barcelona, Spain \\
$^3$ LIGM, Ecole des Ponts, Univ Gustave Eiffel, CNRS, France
$^4$ Tencent AI Lab \\
$^1${\tt\small \{wen.guo,xavier.alameda-pineda\}@inria.fr}, 
$^2${\tt\small \{fmoreno\}@iri.upc.edu}, \\
$^3${\tt\small \{yuming.du,vincent.lepetit\}@enpc.fr},
$^4${\tt\small \{shenxiluc\}@gmail.com}}

\maketitle
\thispagestyle{empty}

\renewcommand{\thefootnote}{\fnsymbol{footnote}}
\footnotetext[1]{Equal contribution.}
\footnotetext[2]{Corresponding author.}
\renewcommand*{\thefootnote}{\arabic{footnote}}

\begin{abstract}
This paper tackles the problem of human motion prediction, consisting in forecasting future body poses from historically observed sequences. State-of-the-art approaches provide good results, however, they rely on deep learning architectures of arbitrary complexity, such as Recurrent Neural Networks~(RNN), Transformers or Graph Convolutional Networks~(GCN), typically requiring multiple training stages and more than 2 million parameters. In this paper, we show that, after combining with a series of standard practices, such as applying Discrete Cosine Transform~(DCT), predicting residual displacement of joints and optimizing velocity as an auxiliary loss, a light-weight network based on multi-layer perceptrons~(MLPs) with only 0.14 million parameters can surpass the state-of-the-art performance.
An exhaustive evaluation on the Human3.6M, AMASS, and 3DPW datasets shows that our method, named \method, consistently outperforms all other approaches. We hope that our simple method could serve as a strong baseline for the community and allow re-thinking of the human motion prediction problem. The code is publicly available at \url{https://github.com/dulucas/siMLPe}. 
\end{abstract}

\section{Introduction}
\label{intro}

Given a sequence of 3D body poses, the task of human motion prediction aims to predict the follow-up of the pose sequence. Forecasting future human motion is at the core of a number of applications, including preventing accidents in autonomous driving~\cite{paden2016survey}, tracking people~\cite{gong2011multi}, or human-robot interaction~\cite{koppula2013anticipating}.

\begin{figure}
    \centering
    \includegraphics[width=1\linewidth]{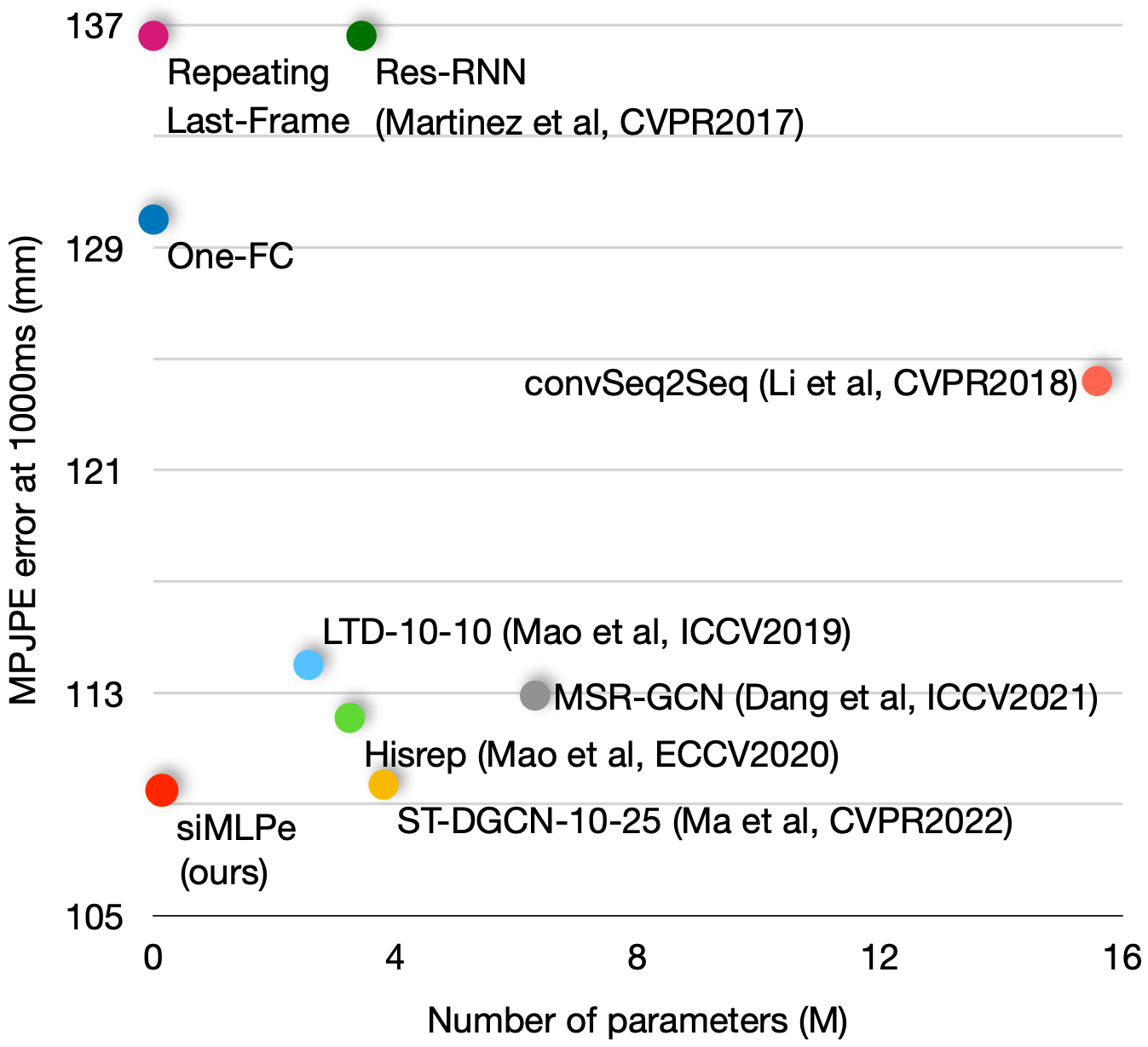}
    \caption{{\bf Comparison of parameter size and performance on the Human3.6M dataset~\cite{ionescu2013human3}.} We report the MPJPE metric in \textit{mm} at 1000 ms as performance on the vertical axis. The closer to bottom-left, the better. Our method (\method, in red) achieves the lowest error with significantly fewer parameters. We also show the performance of two simple methods: `Repeating Last-Frame’ systematically repeats the last input frame as output prediction, and `One-FC’ uses only one single fully connected layer to predict the future motion.}
    \label{fig:teaser}
\end{figure}

Due to the spatio-temporal nature of human motion, the common trend in the literature is to design models that are capable of fusing spatial and temporal information. Traditional approaches mainly relied on hidden Markov models~\cite{brand2000style} or Gaussian process latent variable models~\cite{wang2007gaussian}. However, while these approaches performed well on simple and periodic motion patterns, they dramatically fail under complex motions~\cite{mao2019learning}. In recent years, with the success of deep learning, various methods have been developed based on different types of neural networks that are able to handle sequential data. For example, some works use Recurrent Neural Networks~(RNN)~\cite{martinez2017human} to model the human motion~\cite{fragkiadaki2015recurrent,jain2016structural,martinez2017human,liu2019towards,chiu2019action}, and some more recent works~\cite{mao2019learning, mao2020history, guomulti, ma2022progressively, dang2021msr, li2020dynamic, li2021symbiotic} propose networks based on Graph Convolutional Networks~(GCN)~\cite{mao2019learning}, or trying with Transformers(\cite{aksan2021spatio}) based method~\cite{mao2020history,aksan2021spatio,cai2020learning} to fuse the spatial and temporal information of the motion sequence across human joints and time. However, the architectures of these recent methods are usually not simple and some of them require additional priors, which makes their network difficult to analyze and modify. Thus, a question naturally arises:~ \textit{``Can we tackle the human motion prediction with a simple network?''}

To answer this question, we first tried a naive solution by just repeating the last input pose and using them as the output prediction. As shown in Figure~\ref{fig:teaser}, this naive solution could already achieve reasonable results, which means the last input pose is ``close'' to the future poses. Inspired by this, we further train only one fully connected layer to predict the residual between the future poses and the last input pose and achieves better performance, which shows the potential of a simple network for human motion prediction built on basic layers like the fully connected layer.

Based on the above observations, we go back to the multi-layer perceptrons~(MLPs) and build a simple yet effective network named \method with only three components: fully connected layers, layer normalization~\cite{ba2016layer} and transpose operations. The network architecture is shown in Figure~\ref{fig:pip}. Noticeably,  we found that even commonly used activation layers such as ReLU~\cite{nair2010rectified} are not needed, which makes our network an entirely linear model except for layer normalization. Despite its simplicity, \method achieves strong performance when appropriately combined with three simple practices: applying the Discrete Cosine Transform~(DCT), predicting residual displacement of joints, and optimizing velocity as an auxiliary loss.

\method yields state-of-the-art performance on several standard benchmarks, including Human3.6M~\cite{ionescu2013human3}, AMASS~\cite{mahmood2019amass} and 3DPW~\cite{vonMarcard2018}. In the meantime, \method is lightweight and requires $20\times$ to $60\times$ fewer parameters than previous state-of-the-art approaches. 
A comparison between \method and previous methods can be found in Figure~\ref{fig:teaser}, which shows the Mean Per Joint Position Error~(MPJPE) at $1,000$ms on Human3.6M of different networks versus the network complexity. \method achieves the best performance with high efficiency.

In summary, our contributions are as follows:
\begin{itemize}
    \vspace{-1mm}
    \item We show that human motion prediction can be modeled in a simple way without explicitly fusing spatial and temporal information. As an extreme example, a single fully connected layer can already achieve reasonable performance.
    \vspace{-1mm}
    \item We propose \method, a simple yet effective network for human motion prediction with only three components: fully connected layers, layer normalization, and transpose operation, achieving state-of-the-art performance with far fewer parameters than existing methods on multiple benchmarks such as Human3.6M, AMASS and 3DPW datasets.
\end{itemize}

\section{Related Work}

Human motion prediction is formulated as a sequence-to-sequence task, where past observed motion is taken as input to predict the future motion sequence. Traditional methods explore human motion prediction with nonlinear Markov models~\cite{lehrmann2014efficient}, Gaussian Process dynamical models~\cite{wang2005gaussian}, and Restricted Boltzmann Machine~\cite{taylor2007modeling}. These approaches have shown to be effective to predict simple motions and eventually struggle with complex and long-term motion prediction~\cite{fragkiadaki2015recurrent}. 
With the deep learning era, human motion prediction has achieved great success with the use of deep networks, including Recurrent Neural Networks~(RNNs)~\cite{fragkiadaki2015recurrent,jain2016structural,martinez2017human,liu2019towards,chiu2019action}, 
Graph Convolutional Networks~(GCNs)~\cite{mao2019learning, mao2020history, guomulti, ma2022progressively, dang2021msr, li2020dynamic, li2021symbiotic} and Transformers~\cite{mao2020history,aksan2021spatio,cai2020learning}, which are the main focus of this section.

\subsection{RNN-based human motion prediction}
Due to the inherent sequential structure of human motion, some works address 3D human motion prediction by recurrent models. Fragkiadaki~\etal~\cite{fragkiadaki2015recurrent} propose an encoder-decoder framework to embed human poses and an LSTM to update the latent space and predict future motion.
Jain~\etal~\cite{jain2016structural} manually encode the semantic similarity between different parts of the body and forwards them via structural RNNs. However, these two methods suffer from discontinuity and they are only trained on action-specific models, \ie, a single model is trained for a specific action. 

Martinez~\etal~\cite{martinez2017human} studied multi-actions instead of action-specific models, \ie, train one single model for multiple actions, which  allows the network to exploit regularities across different actions in large-scale datasets. This is widely adopted by most of the subsequent   works. They also introduced a residual connection to model the velocities instead of the absolute value to have more smooth predictions.

Nevertheless, the above-mentioned methods suffer from multiple inherent limitations of RNNs. First, as a sequential model, RNNs are difficult to parallelize during training and inference. Second, the memory constraints prevent RNNs from exploring information from farther frames. Some works alleviate this problem by using RNN variants~\cite{liu2019towards, chiu2019action}, sliding windows~\cite{butepage2017deep, butepage2018anticipating}, convolutional models~\cite{hernandeziccv2019,li2018convolutional} or adversarial training~\cite{gui2018adversarial}, as described in the following sections. But their networks are still complicated and  have a large number of parameters.

\begin{figure*}[t]
    \centering
    \includegraphics[width=0.95\linewidth]{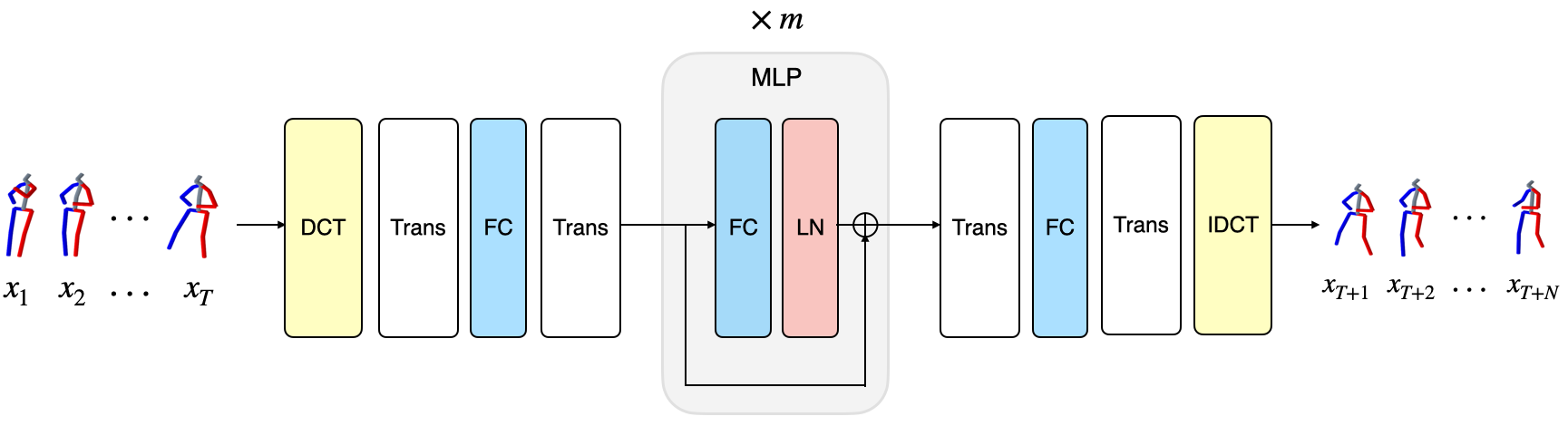}\vspace{2mm}
    \caption{{\bf Overview of our approach \textsc{siMLPe} for human motion prediction.} \textit{FC} denotes a fully connected layer, \textit{LN} denotes layer normalization~\cite{ba2016layer}, and \textit{Trans} represents the transpose operation. \textit{DCT} and \textit{IDCT} represent the discrete cosine transformation and inverse discrete cosine transformations respectively. The MLP blocks (in gray), composing FC and LN, are repeated $m$ times.}
    \label{fig:pip}
    \vspace{-2mm}
\end{figure*}

\subsection{GCN-based human motion prediction}
To better encode the spatial connectivity of human joints, the most recent works usually build the human pose as a graph and adopt Graph Convolutional Networks~(GCNs)~\cite{sperduti1997supervised,kipf2016semi} for human motion prediction.

GCNs were first exploited for human motion prediction in Mao~\etal~\cite{mao2019learning}. They use a stack of blocks consisting of GCNs, nonlinear activation, and batch normalization to encode the spatial dependencies, and leverage discrete cosine transform~(DCT) to encode temporal information.
This work inspired most of the GCN-based motion prediction methods in recent years.
Based on \cite{mao2019learning}, Mao~\etal~\cite{mao2020history} further improved the temporal encoding by cutting the past observations into several sub-sequences and adding an attention mechanism to find similar previous motion sub-sequences in the past with the current observations. 
Thus, the future sequence is computed as a weighted sum of observed sub-sequences. Then, a GCN-based predictor, the same as in \cite{mao2019learning}, is used to encode the spatial dependencies. 

Instead of using DCT transformation to encode the input sequence, \cite{lebailly2020motion} used a multi-scale temporal input embedding, by applying various size convolutional layers for different input sizes to have different receptive fields in the temporal domain. Ma~\etal~\cite{ma2022progressively} proposed two variants of GCNs to extract spatial and temporal features. They built a multi-stage structure where each stage contains an encoder and a decoder, and during the training, the model is trained with intermediate supervision to learn to progressively refine the prediction. \cite{dang2021msr,li2020dynamic,li2021symbiotic} extend the graph of human pose to multi-scale version across the abstraction levels of human pose. 

\subsection{Attention-based human motion prediction}
With the development of transformers~\cite{vaswani2017attention}, some works~\cite{mao2020history,aksan2021spatio,cai2020learning} attempted to deal with this task with an attention mechanism. \cite{mao2020history} used attention to find temporal relations; \cite{aksan2021spatio} also used attention to map not only the temporal dependencies but also the pairwise relation of joints by an architecture combining ``spatial attention'' and ``temporal attention'' in parallel.
\cite{cai2020learning} used a transformer-based architecture along with a progressive-decoding strategy to predict the DCT coefficients of the target joints progressively based on the kinematic tree. In order to guide the predictions, they also built a memory-based dictionary to preserve the global motion patterns in training data. 

In summary, with the development of human motion prediction in recent years, the RNN/GCN/transformer-based architectures are well explored and the results have been significantly improved. Though these methods provide good results, their architectures are becoming more and more complicated and difficult to train.
In this paper, we stick to simple architectures and propose an MLP-based network.
Recently, a concurrent and independent work \cite{bouazizi2022motionmixer} based on ~\cite{tolstikhin2021mlp} also adopts an MLP based network architecture for motion prediction, while our network is much simpler as we do not use the squeeze-and-excitation block\cite{hu2018squeeze} nor the activation layers.
We hope that our simple method would serve as a baseline and let the community rethink the problem of human motion prediction.

\vspace{-2mm}
\section{Our Approach: \method}
\label{sec:method}
In this section, we formulate the problem and present the formulation of the DCT transformation in Section~\ref{sec:data_re}, details of the network architecture in Section~\ref{method:arch} and the losses we use for training in Section~\ref{method:loss}. 
 
Given a sequence of 3D human poses in the past, our goal is to predict the future sequence of poses.
We denote the observed 3D human poses as $\inmotion = [\motion_1^\top, .., \motion_T^\top]^\top \in \mathbb{R}^{T \times C}$, consisting of $T$ consecutive human poses, where the pose at the $t$-th frame $\motion_t$ is represented by a $C$-dimensional vector, \ie\ $\motion_t \in \mathbb{R}^C$. In this work, similar to previous works~\cite{martinez2017human,mao2019learning,mao2020history,ma2022progressively}, $\motion_t$ is the 3D coordinates of joints at $t$-th frame and $C = 3 \times K$, where $K$ is the number of joints. Our task is to predict the future $N$ motion frames $\futuremotion = [\motion_{T+1}^\top, .., \motion_{T+N}^\top]^\top \in \mathbb{R}^{N \times C}$.

\begin{table*}[htbp!]
\caption{\textbf{Results on Human3.6M} for different prediction time steps~(ms). We report the MPJPE error in \textit{mm} and number of parameters (M) for each method. Lower is better. 256 samples are tested for each action. $\dag$ indicates that the results are taken from the paper~\cite{mao2020history}, $\star$ indicated that the results are taken from the paper~\cite{ma2022progressively}. Note that ST-DGCN~\cite{ma2022progressively} use two different models to evaluate their short-/long- term performance, here we report their results of a single model which performs better on long-term for fair comparison. We also show results of two simple baselines: 'Repeating Last-Frame' repeats the last input frame 25 times as output, 'One FC' uses only one single fully connected layer for the prediction.}
\vspace{2mm}
\label{table:tab_res_h36m}
\centering
\rowcolors{3}{gray!15}{white}
\resizebox{0.9\textwidth}{!}{\setlength{\tabcolsep}{3mm}{
\begin{tabular}{l|cccccccc|c}
\toprule
\multirow{1}{*}{}  & \multicolumn{8}{c|}{MPJPE (mm) $\downarrow$} & \multirow{2}{*}{$\#$ Param.(M) $\downarrow$} \\
Time (ms) & 80 & 160 & 320 & 400 & 560 & 720 & 880 & 1000  & \multirow{2}{*}{}\\
\midrule
{\cellcolor{white}Repeating Last-Frame} & 23.8 & 44.4 & 76.1 & 88.2 & 107.4 & 121.6 & 131.6 & 136.6 & 0 \\
{\cellcolor{white}One FC} & 14.0 & 33.2 & 68.0 & 81.5 & 101.7 & 115.1 & 124.8 & 130.0 & 0.003 \\
\midrule
{\cellcolor{white}Res-RNN $\dag$ \cite{martinez2017human}}& 25.0 & 46.2 & 77.0 & 88.3 & 106.3 & 119.4 & 130.0 & 136.6  & 3.44 \\ 
{\cellcolor{white}convSeq2Seq $\dag$ \cite{li2018convolutional}} & 16.6 & 33.3 & 61.4 & 72.7 & 90.7 & 104.7 & 116.7 & 124.2 & 15.58 \\ 
{\cellcolor{white}LTD-50-25 $\dag$ \cite{mao2019learning}} & 12.2 & 25.4 & 50.7 & 61.5 & 79.6 & 93.6 & 105.2 & 112.4 & 2.56 \\ 
{\cellcolor{white}LTD-10-10 $\dag$ \cite{mao2019learning}}  & 11.2 & 23.4 & 47.9 & 58.9 & 78.3 & 93.3 & 106.0 & 114.0 & 2.55\\ 
{\cellcolor{white}Hisrep $\dag$ \cite{mao2020history}} & 10.4 & 22.6 & 47.1 & 58.3 & 77.3 & 91.8 & 104.1 & 112.1  & 3.24\\ 
{\cellcolor{white}MSR-GCN $\star$ \cite{dang2021msr}}  & 11.3 & 24.3 & 50.8 & 61.9 & 80.0 & - & - & 112.9 & 6.30\\ 
{\cellcolor{white}ST-DGCN-10-25 $\star$ \cite{ma2022progressively}} & 10.6 & 23.1 & 47.1 & 57.9 & 76.3 & 90.7 & 102.4 & 109.7 & 3.80  \\
\midrule
\cellcolor{white}\method~(Ours)  & \bf 9.6 & \bf 21.7 & \bf 46.3 & \bf 57.3 & \bf 75.7 & \bf 90.1 & \bf 101.8 & \bf 109.4 & \bf 0.14\\

\bottomrule
\end{tabular}}}
\end{table*}
\setlength{\tabcolsep}{1.4pt}
\begin{table*}[ht]
\caption{\textbf{Action-wise results on Human3.6M} for different prediction time steps~(ms). Lower is better. 256 samples are tested for each action. $\dag$ indicates that the results are taken from the paper~\cite{mao2020history}, $\star$ indicates that the results are taken from the paper~\cite{ma2022progressively}.}
\vspace{2mm}
\centering
\rowcolors{1}{gray!15}{white}
\resizebox{0.99\textwidth}{!}{\setlength{\tabcolsep}{1.5mm}{
\begin{tabular}{l|cccc|cccc|cccc|cccc} 
\toprule
Action & \multicolumn{4}{c|}{walking}  & \multicolumn{4}{c|}{eating} & \multicolumn{4}{c|}{smoking} & \multicolumn{4}{c}{discussion} \\ 
\midrule
Time~(ms)  & 80 & 400 & 560 & 1000 & 80 & 400 & 560 & 1000 & 80 & 400 & 560 & 1000 & 80 & 400 & 560 & 1000 \\ 
\midrule

{\cellcolor{white}Res-RNN $\dag$ \cite{martinez2017human}} &  23.2 & 66.1 & 71.6 & 79.1 & 16.8 &  61.7 & 74.9 & 98.0 & 18.9  & 65.4 & 78.1 & 102.1& 25.7 & 91.3 & 109.5  & 131.8 \\ 
{\cellcolor{white}convSeq2Seq $\dag$ \cite{li2018convolutional}} & 17.7  & 63.6 &  72.2& 82.3& 11.0  & 48.4  & 61.3& 87.1& 11.6 &  48.9  & 60.0& 81.7& 17.1  & 77.6  & 98.1&129.3 \\
{\cellcolor{white}LTD-50-25 $\dag$ \cite{mao2019learning}} & 12.3  & 44.4  & 50.7 & 60.3 & 7.8 &  38.6 &51.5 & 75.8 & 8.2 &  39.5  & 50.5 &72.1 & 11.9 & 68.1  &88.9 & 118.5\\
{\cellcolor{white}LTD-10-10 $\dag$ \cite{mao2019learning}}  & 11.1  & 42.9  & 53.1 & 70.7 & 7.0 &  37.3 & 51.1 & 78.6 & 7.5  & 37.5  & 49.4 & 71.8 & 10.8 & 65.8  &88.1 & 121.6\\
{\cellcolor{white}Hisrep $\dag$ \cite{mao2020history}} & 10.0 & 39.8  & 47.4 & 58.1 & 6.4  & 36.2  & 50.0 & 75.7 & 7.0 & 36.4  & 47.6  & 69.5 & 10.2 & 65.4  & 86.6 & 119.8\\ 
{\cellcolor{white}MSR-GCN $\star$ \cite{dang2021msr}}   &  10.8  &  42.4  &  53.3  &  63.7  &  6.9  & 36.0  & 50.8 & 75.4 & 7.5 &  37.5  & 50.5 & 72.1 & 10.4 & 65.0  & 87.0 & 116.8\\ 
{\cellcolor{white}ST-DGCN-10-25 $\star$ \cite{ma2022progressively}} & 11.2	 & 42.8 & 	49.6 & 	58.9  &  6.5 & 	36.8 & 	50.0 & 	74.9  &    7.3 & 	37.5 & 	48.8	 & 69.9 &  10.2 & 	64.4 & 	86.1 & 	116.9 \\
\midrule
\cellcolor{white}\method~(Ours)  &\textbf{ 9.9}  & \textbf{  39.6 } &   \textbf{46.8}  &  \textbf{55.7 }  &  \textbf{5.9}  &  \textbf{36.1}  &  \textbf{ 49.6}  &  \textbf{ 74.5}   & \textbf{ 6.5}  &   \textbf{36.3}  &   \textbf{47.2}  &  \textbf{69.3}   & \textbf{ 9.4 } &  \textbf{ 64.3}  &  \textbf{85.7}  &   \textbf{116.3} \\

\midrule
Action & \multicolumn{4}{c|}{directions}  & \multicolumn{4}{c|}{greeting} & \multicolumn{4}{c|}{phoning} & \multicolumn{4}{c}{posing} \\ 
\midrule
Time~(ms)  & 80 & 400 & 560 & 1000 & 80 & 400 & 560 & 1000 & 80 & 400 & 560 & 1000 & 80 & 400 & 560 & 1000 \\  
\midrule

{\cellcolor{white}Res-RNN $\dag$ \cite{martinez2017human}}& 21.6 & 84.1 & 101.1 & 129.1 & 31.2 & 108.8 & 126.1 & 153.9 & 21.1 & 76.4 & 94.0 & 126.4 & 29.3 & 114.3 & 140.3 & 183.2\\ 
{\cellcolor{white}convSeq2Seq $\dag$ \cite{li2018convolutional}} & 13.5 & 69.7 & 86.6 & 115.8 & 22.0 & 96.0 & 116.9 & 147.3 & 13.5 & 59.9 & 77.1 & 114.0 & 16.9 & 92.9 & 122.5 & 187.4 \\ 
{\cellcolor{white}LTD-50-25 $\dag$ \cite{mao2019learning}} & 8.8 & 58.0 & 74.2 & 105.5 & 16.2 & 82.6 & 104.8 & 136.8 & 9.8 & 50.8 & 68.8 & 105.1 & 12.2 & 79.9 & 110.2 & 174.8 \\
{\cellcolor{white}LTD-10-10 $\dag$ \cite{mao2019learning}}  & 8.0 & 54.9 & 76.1 & 108.8 & 14.8 & 79.7 & 104.3 & 140.2 & 9.3 & 49.7 & 68.7 & 105.1 & 10.9 & 75.9 & 109.9 & 171.7  \\ 
{\cellcolor{white}Hisrep $\dag$ \cite{mao2020history}} & 7.4 & 56.5 & 73.9 & 106.5 & 13.7 & 78.1 & 101.9 & 138.8 & 8.6 & 49.2 & 67.4 & 105.0 & 10.2 & 75.8 & 107.6 & 178.2 \\ 
{\cellcolor{white}MSR-GCN $\star$ \cite{dang2021msr}} & 7.7 & 56.2 & 75.8 & 105.9 & 15.1  & 85.4 & 106.3 & \textbf{136.3} & 9.1  & 49.8 & 67.9 & 104.7 & 10.3 &  75.9 & 112.5 & 176.5\\ 
{\cellcolor{white}ST-DGCN-10-25 $\star$ \cite{ma2022progressively}}  & 7.5 & 	56.0 & 	73.3 & 	\textbf{105.9 } & 14.0 & 77.3 & 	100.2 & 136.4  & 8.7 & 	48.8 & 	66.5	 & \textbf{102.7 } & 10.2 & 	73.3 & 	102.8 & \textbf{167.0}\\
\midrule
\cellcolor{white}\method~(Ours)  & \textbf{6.5}  &   \textbf{55.8}  &   \textbf{73.1}  &   106.7 &  \textbf{12.4 } &   \textbf{77.3}  &   \textbf{99.8}  &  137.5   & \textbf{ 8.1}  &  \textbf{ 48.6}  &   \textbf{66.3}  &  103.3   &  \textbf{8.8}  &  \textbf{73.8}  &  \textbf{103.4}  &  168.7 \\

\midrule
Action  & \multicolumn{4}{c|}{purchases} & \multicolumn{4}{c|}{sitting}  & \multicolumn{4}{c|}{sittingdown}  & \multicolumn{4}{c}{takingphoto}\\ 
\midrule
Time~(ms)  & 80 & 400 & 560 & 1000 & 80 & 400 & 560 & 1000 & 80 & 400 & 560 & 1000 & 80 & 400 & 560 & 1000 \\ 
\midrule

{\cellcolor{white}Res-RNN $\dag$ \cite{martinez2017human}} & 28.7 & 100.7 & 122.1 & 154.0 & 23.8 & 91.2 & 113.7 & 152.6 & 31.7 & 112.0 & 138.8 & 187.4 & 21.9 & 87.6 & 110.6 & 153.9 \\ 
{\cellcolor{white}convSeq2Seq $\dag$ \cite{li2018convolutional}} & 20.3 & 89.9 & 111.3 & 151.5 & 13.5 & 63.1 & 82.4 & 120.7  & 20.7 & 82.7 & 106.5 & 150.3 & 12.7 & 63.6 & 84.4 & 128.1  \\ 
{\cellcolor{white}LTD-50-25 $\dag$ \cite{mao2019learning}} & 15.2 & 78.1 & 99.2 & 134.9 & 10.4 & 58.3 & 79.2 & 118.7 & 17.1 & 76.4 & 100.2 & 143.8 & 9.6 & 54.3  & 75.3 & 118.8  \\
{\cellcolor{white}LTD-10-10 $\dag$ \cite{mao2019learning}}  & 13.9 & 75.9 & 99.4 & 135.9 & 9.8 & 55.9 & 78.5 & 118.8  & 15.6  &  71.7  & 96.2 & 142.2 & 8.9  &  51.7  & 72.5 & 116.3\\ 
{\cellcolor{white}Hisrep $\dag$ \cite{mao2020history}} & 13.0 & 73.9 & 95.6 & 134.2 & 9.3 & 56.0 & 76.4 & 115.9 & 14.9  &  72.0  & 97.0 & 143.6 & 8.3  &  51.5  & 72.1 & 115.9 \\ 
{\cellcolor{white}MSR-GCN $\star$ \cite{dang2021msr}}  & 13.3 & 77.8 & 99.2 & 134.5 & 9.8 & 55.5 & 77.6 & 115.9 & 15.4  & 73.8 & 102.4 & 149.4 & 8.9 & 54.4 & 77.7 & 121.9 \\ 
{\cellcolor{white}ST-DGCN-10-25 $\star$ \cite{ma2022progressively}} & 13.2 & 	74.0 & 	95.7 & 	\textbf{132.1}  &  9.1 & 	54.6 & 	75.1 & 	114.8  &  14.7 & \textbf{	70.0} & 	\textbf{94.4} & 	\textbf{139.0} &   8.2 & 	\textbf{50.2} & 	\textbf{70.5} & 	112.9 \\
\midrule
\cellcolor{white}\method~(Ours)  & \textbf{11.7}  &   \textbf{72.4}  &   \textbf{93.8}  &   132.5   &  \textbf{8.6}  &   \textbf{55.2}  &  \textbf{75.4 } &  \textbf{ 114.1}   &  \textbf{13.6 } & 70.8  &   95.7  &   142.4  &  \textbf{7.8 } &   50.8  &   71.0 &  \textbf{ 112.8}\\

\midrule
Action & \multicolumn{4}{c|}{waiting}  & \multicolumn{4}{c|}{walkingdog} & \multicolumn{4}{c|}{walkingtogether} & \multicolumn{4}{c}{average} \\ 
\midrule
Time~(ms)  & 80 & 400 & 560 & 1000 & 80 & 400 & 560 & 1000 & 80 & 400 & 560 & 1000 & 80 & 400 & 560 & 1000 \\ 
\midrule

{\cellcolor{white}Res-RNN $\dag$ \cite{martinez2017human}}& 23.8 & 87.7 & 105.4 & 135.4 & 36.4 & 110.6 & 128.7 & 164.5  & 20.4 & 67.3 & 80.2 & 98.2 & 25.0 & 88.3 & 106.3 & 136.6\\ 
{\cellcolor{white}convSeq2Seq $\dag$ \cite{li2018convolutional}}& 14.6 & 68.7 & 87.3 & 117.7 & 27.7 & 103.3 & 122.4 & 162.4 & 15.3 & 61.2 & 72.0 & 87.4 & 16.6 & 72.7 & 90.7 & 124.2\\ 
{\cellcolor{white}LTD-50-25 $\dag$ \cite{mao2019learning}} &  10.4 & 59.2   & 77.2 & 108.3 & 22.8 & 88.7  & 107.8 & 156.4 & 10.3 & 46.3  & 56.0 & 65.7 & 12.2 & 61.5  & 79.6 & 112.4\\
{\cellcolor{white}LTD-10-10 $\dag$ \cite{mao2019learning}}  & 9.2  &  54.4  & 73.4 & 107.5 & 20.9  &  86.6  & 109.7 & 150.1 & 9.6  &  44.0  & 55.7 & 69.8 & 11.2  &  58.9 &  78.3 & 114.0 \\ 
{\cellcolor{white}Hisrep $\dag$ \cite{mao2020history}}  & 8.7  &  54.9  & 74.5 & 108.2 & 20.1  &  86.3  & 108.2 & 146.9 & 8.9  &  41.9  & 52.7 & 64.9 & 10.4  &  58.3 & 77.3 & 112.1 \\ 

{\cellcolor{white}MSR-GCN $\star$ \cite{dang2021msr}}  & 10.4  & 62.4 & 74.8 & 105.5 & 24.9 & 112.9 & 107.7 & 145.7 & 9.2 & 43.2 & 56.2 & 69.5 & 11.3 & 61.9 & 80.0 & 112.9\\ 
{\cellcolor{white}ST-DGCN-10-25 $\star$ \cite{ma2022progressively}} & 8.7 & 53.6 & 	71.6 & 	\textbf{103.7}  &  20.4 & 	84.6 & 	105.7 & 145.9  &  8.9 &	43.8 & 	54.4 & 	64.6  &  10.6 & 	57.9 & 	76.3 & 	109.7 \\
\midrule
\cellcolor{white}\method~(Ours)  &\textbf{7.8}  &  \textbf{53.2}  &   \textbf{71.6} &   104.6  &  \textbf{18.2} &   \textbf{83.6} &  \textbf{105.6} &  \textbf{141.2} &  \textbf{8.4} &   \textbf{41.2} &   \textbf{50.8}  &  \textbf{61.5}  &   \textbf{9.6} & \textbf{57.3} & \textbf{75.7} & \textbf{109.4}\\

\bottomrule
\end{tabular}
}}
\vspace{-4mm}
\label{table:tab_res_h36m_act_wise}
\end{table*}

\subsection{Discrete Cosine Transform~(DCT)}
\label{sec:data_re}

We adopt the DCT transformation to encode temporal information, which is proven to be beneficial for human motion prediction~\cite{mao2019learning,mao2020history,ma2022progressively}. More precisely, given an input motion sequence of $T$ frames, the DCT matrix $\mathbf{D} \in \mathbb{R}^{T \times T}$ can be calculated as: 
\begin{align}
    \mathbf{D}_{i,j} = \sqrt{\frac{2}{T}} \frac{1}{\sqrt{1+\delta_{i,0}}}\cos\left(\frac{\pi}{2T}(2j+1)i \right) \> ,
\end{align}
where $\delta_{i,j}$ denotes the~\emph{Kronecker} delta:
\begin{align}
  \delta_{i,j} = \begin{cases}
  1 & \text{if}\ i=j\\
  0 & \text{if}\ i\neq j \> .
  \end{cases}
\end{align}

The transformed input is $\mathcal{D}(\inmotion) = \mathbf{D} \inmotion $. We apply the Inverse Discrete Cosine Transform~(IDCT) to transform the output of the network back to the original pose representation, denoted as $\mathcal{D}^{-1}$ and the inverse of $\mathbf{D}$.

\subsection{Network architecture}
\label{method:arch}

Figure~\ref{fig:pip} shows the architecture of our network. Our network only contains three components: fully connected layers, transpose operation, and layer normalization~\cite{ba2016layer}. For  all the fully connected layers, their input dimension is equal to their output dimension.

Formally, given an input sequence of 3D human poses $\inmotion = [\motion_1^\top, .., \motion_T^\top]^\top \in \mathbb{R}^{T \times C}$, our network predicts a sequence of future poses $\predmotion = [{\motion'}_{T+1}^\top, .., {\motion'}_{T+N}^\top]^\top \in \mathbb{R}^{N \times C}$:
\begin{align}
    \predmotion = \mathcal{D}^{-1}(\mathcal{F}(\mathcal{D}(\inmotion))) \> ,
\end{align}
where $\mathcal{F}$ denotes our network.

After the DCT transformation, we apply one fully connected layer to operate only on the spatial dimension of the transformed motion sequence $\mathcal{D}(\inmotion) \in \mathbb{R}^{T \times C}$:
\begin{align}
     \mathbf{z}^0 = \mathcal{D}(\inmotion) \mathbf{W}_0 + \mathbf{b}_0 \> ,
\end{align}
where $\mathbf{z}^0 \in \mathbb{R}^{T \times C}$ is the output of the fully connected layer, $\mathbf{W}_0 \in \mathbb{R}^{C \times C}$ and $\mathbf{b}_0 \in \mathbb{R}^{C}$ represent the learnable parameters of the fully connected layer. In practice, this is equivalent to applying a transpose operation with a fully connected layer, and then transposing  back the output feature, as shown in Figure~\ref{fig:pip}.

Then, a series of $m$ blocks are introduced to only operate on the temporal dimension, \ie, only to merge information across frames. Each block consists of a fully connected layer followed by layer normalization, formally:
\begin{align}
     \mathbf{z}^i = \mathbf{z}^{i-1} + \text{LN}(\mathbf{W}_i \mathbf{z}^{i-1} + \mathbf{b}_i) \> ,
\end{align}
where $\mathbf{z}^i \in \mathbb{R}^{T \times C}, i \in [1,..,m]$ denotes the output of the $i$-th MLP block, $\text{LN}$ denotes the layer normalization operation, and $\mathbf{W}_i \in \mathbb{R}^{T \times T}$ and $\mathbf{b}_i \in \mathbb{R}^{T}$ are the learnable parameters of the fully connected layer in the $i$-th MLP block.

Finally, similar to the first fully connected layer, we add another fully connected layer after the MLP blocks to operate only on the spatial dimension of the feature, and then apply IDCT transformation to obtain the prediction:
\begin{align}
     \predmotion = \mathcal{D}^{-1}(\mathbf{z}' \mathbf{W}_{m+1} + \mathbf{b}_{m+1}) \> ,
\end{align}
where $\mathbf{W}_{m+1}$ and $\mathbf{b}_{m+1}$ are the learnable parameters of the last fully connected layer.

Note that the lengths $T$ and $N$ do not need to be equal. When $T > N$, we only take the $N$ first frames of the prediction, and in the case of $T < N$, we could pad our input sequence to $N$ by repeating the last frame, as done in \cite{mao2019learning, mao2020history}.

\subsection{Losses}
\label{method:loss}

As mention in the Section~\ref{intro} and shown in the Figure~\ref{fig:teaser}, the last input pose is ``close'' to the future poses. Inspired by this observation, instead of predicting the absolute 3D poses from scratch, we let our network predict the residual between the future pose $\motion_{T+t}$ and the last input pose $x_T$. As we will show in Section~\ref{sec:abl}, this eases the learning and improves performance.

\vspace{2mm} 
\paragraph{Objective function.} Our objective function $\mathcal{L}$ includes two terms $\mathcal{L}_{re}$ and $\mathcal{L}_{v}$:
\begin{align}
    \mathcal{L} &= \mathcal{L}_{re}  + \mathcal{L}_{v} \> .
\end{align}
$\mathcal{L}_{re}$ aims to minimize the $\mathcal{L}_2$-norm between the predicted motion $\predmotion$ and ground-truth one $\futuremotion$:
\begin{align}
    \mathcal{L}_{re} &= \mathcal{L}_{2}(\predmotion, \futuremotion) \>.
\end{align}

$\mathcal{L}_{v}$ aims to minimize the $\mathcal{L}_2$-norm between the velocity of the predicted motion $\predspeed$ and the ground truth one $\futurespeed$:
\begin{align}
    \mathcal{L}_{v} &= \mathcal{L}_{2}(\predspeed, \futurespeed) \> ,
\end{align}
where $\futurespeed = [\speed_{T+1}^\top, .., \speed_{T+N}^\top]^\top \in \mathbb{R}^{N \times C}$, $\speed_t$ represents the velocity at frame $t$ and is computed as the time difference: $\speed_t = \motion_{t+1} - \motion_t$. We provide a full analysis of the loss terms in Section~\ref{sec:abl}.

\begin{table*}[htbp!]
\caption{\textbf{Results on AMASS and 3DPW} for different prediction time steps~(ms). We report the MPJPE error in \textit{mm}. Lower is better. The model is trained on the AMASS dataset. The results of the previous methods are taken from~\cite{mao2020history}.}\vspace{2mm}
\label{table:tab_res_amass}
\centering
\rowcolors{3}{gray!15}{white}
\resizebox{0.99\textwidth}{!}{\setlength{\tabcolsep}{1.6mm}{
\begin{tabular}{l|cccccccc|cccccccc}
\toprule
Dataset & \multicolumn{8}{c|}{AMASS-BMLrub} & \multicolumn{8}{c}{3DPW} \\

Time (ms) & 80 & 160 & 320 & 400 & 560 & 720 & 880 & 1000 & 80 & 160 & 320 & 400 & 560 & 720 & 880 & 1000\\
\midrule
{\cellcolor{white}convSeq2Seq} \cite{li2018convolutional}  & 20.6 & 36.9 & 59.7 & 67.6 & 79.0 & 87.0 & 91.5 & 93.5 & 18.8 & 32.9 & 52.0 & 58.8 & 69.4 & 77.0 & 83.6 & 87.8\\ 
{\cellcolor{white}LTD-10-10} \cite{mao2019learning} & \bf 10.3 & \bf 19.3 & 36.6 & 44.6 & 61.5 & 75.9 & 86.2 & 91.2 & \bf 12.0 & \bf 22.0 & 38.9 & 46.2 & 59.1 & 69.1 & 76.5 & 81.1\\ 
{\cellcolor{white}LTD-10-25} \cite{mao2019learning} & 11.0 & 20.7 & 37.8 & 45.3 & 57.2 & 65.7 & 71.3 & 75.2 & 12.6 & 23.2 & 39.7 & 46.6 & 57.9 & 65.8 & 71.5 & 75.5\\
{\cellcolor{white}Hisrep} \cite{mao2020history} & 11.3 & 20.7 & 35.7 & 42.0 & 51.7 & 58.6 & 63.4 & 67.2 & 12.6 & 23.1 & 39.0 & 45.4 & 56.0 & 63.6 & 69.7 & 73.7 \\ 

\midrule
\cellcolor{white}\method~(Ours) & 10.8 & 19.6 & \bf 34.3 & \bf 40.5 & \bf 50.5 & \bf 57.3 & \bf 62.4 & \bf 65.7 & 12.1 & 22.1 & \bf 38.1 & \bf 44.5 & \bf 54.9 & \bf 62.4 & \bf 68.2 & \bf 72.2 \\
\bottomrule
\end{tabular}}}

\end{table*}
\vspace{3mm}
\setlength{\tabcolsep}{1.4pt}
\begin{table*}[htbp!]
\caption{\textbf{Average results for different prediction time periods on Human3.6M and AMASS.} These results are obtained following the evaluation method of STS-GCN~\cite{sofianos2021space} and STG-GCN \cite{zhong2022spatio}, instead of the standard evaluation protocol adopted in \cite{mao2019learning,mao2020history,ma2022progressively}.
}\vspace{2mm}
\label{table:tab_res_h36m_avg}
\centering
\rowcolors{3}{gray!15}{white}
\resizebox{0.99\textwidth}{!}{\setlength{\tabcolsep}{1.8mm}{
\begin{tabular}{l|cccccccc|cccccccc}
\toprule
Dataset  & \multicolumn{8}{c|}{Human3.6M} & \multicolumn{8}{c}{AMASS-BMLrub} \\
Time (ms) & 80 & 160 & 320 & 400 & 560 & 720 & 880 & 1000 & 80 & 160 & 320 & 400 & 560 & 720 & 880 & 1000 \\
\midrule
{\cellcolor{white}STS-GCN~\cite{sofianos2021space}}  & 10.1 & 17.1 & 33.1 & 38.3 & 50.8 & 60.1 & 68.9 & 75.6  &10.0 & 12.5 & 21.8 & 24.5 & 31.9 & 38.1 & 42.7 & 45.5 \\ 
{\cellcolor{white}STG-GCN \cite{zhong2022spatio}} & 10.1  & 16.9 & 32.5 & 38.5 & 50.0 & - & - & 72.9  &10.0 & 11.9 & 20.1 & 24.0 & 30.4 &  - & - & 43.1 \\ 
\midrule
\cellcolor{white}\method~(Ours)  & \bf 4.5 & \bf 9.8 & \bf 22.0 & \bf 28.1 & \bf 39.3 & \bf 49.2 & \bf 57.8 & \bf 63.7 & \bf 6.1 & \bf 10.8 & \bf 19.1 & \bf 22.8 & \bf 29.5 & \bf 35.1 & \bf 39.7 & \bf 42.7 \\

\bottomrule
\end{tabular}}}
\vspace{-4mm}
\end{table*}
\setlength{\tabcolsep}{1.4pt}

\section{Experiments}
In this section, we present our experimental details and results. We introduce the datasets and evaluation metric in Section~\ref{sec:datasets}, the implementation details in Section~\ref{method:detail}, and the quantitative/qualitative results in Section~\ref{sec:experimental}. An exhaustive ablation analysis is provided in Section~\ref{sec:abl}.

\subsection{Datasets and evaluation metric}
\label{sec:datasets}

\paragraph{Human3.6M dataset~\cite{ionescu2013human3}.} Human3.6M contains 7 actors performing 15 actions, and 32 joints are labeled for each pose. We follow the same testing protocols of \cite{mao2020history} and use \textit{S5} as the test set, \textit{S11} as the validation set, and the others as the train set.
Previous works use different test sampling strategies, including 8 samples per action~\cite{martinez2017human,mao2019learning}, 256 samples per action~\cite{mao2020history} or all samples in the test set~\cite{dang2021msr}. As 8 samples are too little and taking all testing samples could not balance different actions with different sequence lengths, we thus take 256 samples per action for testing, and evaluate on 22 joints as in~\cite{martinez2017human, mao2019learning, mao2020history, ma2022progressively}. 

\setlength{\tabcolsep}{1.4pt}

\begin{table*}[t!]
\caption{\textbf{Ablation of the number of MLP blocks} on Human3.6M.
}\vspace{2mm}
\label{table:tab_abla_layers}
\centering
\rowcolors{3}{gray!15}{white}
\resizebox{0.8\textwidth}{!}{\setlength{\tabcolsep}{4mm}{
\begin{tabular}{l|c|cccccccc}
\toprule
\multirow{2}{*}{Nb. Blocks}  & \multirow{2}{*}{$\#$ Param.(M) $\downarrow$} & \multicolumn{8}{c}{MPJPE (mm) $\downarrow$}  \\
 & & 80 & 160 & 320 & 400 & 560 & 720 & 880 & 1000 \\
\midrule
\cellcolor{white}1 & 0.012 & 12.7 & 28.5 & 59.7 & 72.1 & 93.6 & 107.0 & 116.8 & 123.6\\
\cellcolor{white}2 & 0.014 & 10.9 & 24.9 & 52.3 & 64.0 & 83.2 & 97.3 & 108.4 & 115.4 \\
\cellcolor{white}6 & 0.025 & 10.2 & 23.1 & 48.8 & 60.1 & 79.0 & 93.3 & 105.1 & 112.6\\
\cellcolor{white}12 & 0.041 & 9.9 & 22.4 & 47.2 & 58.3 & 77.1 & 91.5 & 103.3 & 110.9 \\
\cellcolor{white}24 & 0.073 & 9.7 & 22.0 & 46.8 & 57.7 & 76.4 & 90.8 & 102.6 & 110.3\\
\cellcolor{white}48 (Ours)  & 0.138 & \bf 9.6 & \bf 21.7 & \bf 46.3 & \bf 57.3 & \bf 75.7 & \bf 90.1 & \bf 101.8 & \bf 109.4 \\

\cellcolor{white}64 & 0.180 & 9.6 & 21.8 & 46.5 & 57.5 & 76.0 & 90.1 & 101.9 & 109.7 \\
\cellcolor{white}96 & 0.266 & 9.7 & 21.9 & 46.7 & 57.8 & 76.3 & 90.5 & 102.1 & 109.8\\

\bottomrule
\end{tabular}}}
\vspace{-4mm}
\end{table*}
\setlength{\tabcolsep}{1.4pt}

\begin{table}[t!]
\vspace{2mm}
\caption{\textbf{Ablation of different components of our network} on Human3.6M. \textit{'LN'} denotes the layer normalization. \textit{'DCT'} denotes the DCT transformation. \textit{'Spa. only'} means that all FC layers are on spatial dimensions (w/o transpose operations before/after MLP blocs). \textit{'Temp. only'} means that all FC layers are on temporal dimensions (w/o any transpose operations).}\vspace{2mm} 
\label{table:tab_abla_net}
\centering
\rowcolors{2}{gray!15}{white}
\resizebox{0.48\textwidth}{!}{\setlength{\tabcolsep}{1.1mm}{
\begin{tabular}{l|cccccccc}
\toprule
Ablation & 80 & 160 & 320 & 400 & 560 & 720 & 880 & 1000 \\

\midrule
\cellcolor{white}Spa. only, w/o LN  & 23.7 & 44.0 & 75.5 & 87.6 & 106.3 & 120.4 & 130.5 & 135.6 \\
\cellcolor{white}Spa. only & 23.8 & 43.0 & 73.4 & 85.2 & 102.0 & 116.3 & 125.3 & 131.9 \\
\cellcolor{white}Temp. only & 9.9 & 22.4 & 47.2 & 58.4 & 77.2 & 91.1 & 102.8 & 110.5 \\
\cellcolor{white}w/o LN  & 12.7 & 29.0 & 62.3 & 76.2 & 97.4 & 111.6 & 121.6 & 127.3 \\
\cellcolor{white}w/o DCT & 9.9 & 22.4 & 47.3 & 58.4 & 76.9 & 91.2 & 102.8 & 110.5 \\

\midrule
\cellcolor{white}\method (ours) & \bf 9.6 & \bf 21.7 & \bf 46.3 & \bf 57.3 & \bf 75.7 & \bf 90.1 & \bf 101.8 & \bf 109.4\\

\bottomrule
\end{tabular}}}
\vspace{-2mm}
\end{table}
\setlength{\tabcolsep}{1.4pt}
\begin{table}[t!]
\caption{\textbf{Ablation of data augmentation} on Human3.6M. We only use front-back flip as our data
augmentation, \ie, we randomly invert the motion sequence
during the training.}
\vspace{2mm}
\label{table:tab_abla_aug}
\centering
\resizebox{0.48\textwidth}{!}{\setlength{\tabcolsep}{1.6mm}{
\begin{tabular}{l|cccccccc}
\toprule
 & 80 & 160 & 320 & 400 & 560 & 720 & 880 & 1000 \\
\midrule
w/o aug & 10.0 &  22.6 &  48.3 &  59.7 &  78.2 &  92.0 &  103.4 &  110.8 \\
\rowcolor{gray!15} \cellcolor{white}w aug & \bf 9.6 & \bf 21.7 & \bf 46.3 & \bf 57.3 & \bf 75.7 & \bf 90.1 & \bf 101.8 & \bf 109.4 \\
\bottomrule
\end{tabular}}}
\vspace{-4mm}
\end{table}

\vspace{2mm} 
\paragraph{AMASS dataset~\cite{mahmood2019amass}.} AMASS is a collection of multiple Mocap datasets~\cite{ACCAD,mahmood2019amass,DanceDB:Aristidou:2019,BMLhandball,BMLrub,cmuWEB,dfaust:CVPR:2017,Eyes_Japan,ghorbani2020movi,chatzitofis2020human4d,sigal2010humaneva,mandery2015kit,MoSh_lopermahmoodetal2014,MPI_HDM05,PosePrior_Akhter:CVPR:2015,SFU,TCD_hands,TotalCapture_Trumble}
unified by SMPL parameterization~\cite{loper2015smpl}. We follow \cite{mao2020history} to use AMASS-BMLrub~\cite{BMLrub} as the test set and split the rest of the AMASS dataset into training and validation sets. The model is evaluated on 18 joints as in~\cite{mao2020history}.

\vspace{2mm} 
\paragraph{3DPW dataset~\cite{vonMarcard2018}.} 3DPW is a dataset including indoor and outdoor scenes. A pose is represented by 26 joints, but we follow \cite{mao2020history} and evaluate 18 joints using the model trained on AMASS to evaluate generalization.

\vspace{2mm}  
\paragraph{Evaluation metric.} We report the Mean Per Joint Position Error~(MPJPE) on 3D joint coordinates, which is the most widely used metric for evaluating 3D pose errors. This metric calculates the average L2-norm across different joints between the prediction and ground-truth. Similar to previous works~\cite{mao2019learning,mao2020history,dang2021msr, ma2022progressively}, we ignore the global rotation and translation of the poses and keep the sampling rate as 25 frames per second (FPS) for all datasets.

\subsection{Implementation details}
\label{method:detail}

In practice, we set the input length $T = 50$, the output length $N = 10$ on Human3.6M dataset and $N = 25$ on AMASS dataset and 3DPW dataset.
During testing, we apply our model in an auto-regressive manner to generate motion for longer periods. The feature dimension $C = 3 \times K$, where $K$ is the number of joints, $K = 22$ for Human3.6M and $K = 18$ for AMASS and 3DPW. 

To train our network, we set the batch size to 256 and use the Adam optimizer~\cite{kingma2014adam}. The memory consumed by our network is about $1.5GB$ during the training. 
All our experiments are conducted using the Pytorch~\cite{paszke2019pytorch} framework on a single NVIDIA RTX 2080Ti graphics card. 
We train our network on the Human3.6M dataset for 35k iterations, the learning rate starts from 0.0003 at the beginning and drops to 0.00001 after 30k steps. The training takes $\sim$30 minutes. 
For AMASS dataset, we train our network for 115k iterations. The learning rate starts from 0.0003 at the beginning and drops to 0.00001 after 100k steps. The training takes $\sim$2 hours. 
During training, we only use the front-back flip as  data augmentation, which randomly inverts the motion sequence during the training.

\subsection{Quantitative and qualitative results}
\label{sec:experimental}
In this section, we compare our approach to existing state-of-the-art methods on different datasets. We report MPJPE in \textit{mm} at different prediction time steps up to 1000ms. 

\vspace{2mm}
\paragraph{Human3.6M dataset.} In Table~\ref{table:tab_res_h36m}, we compare our method with other state-of-the-art methods on the Human3.6M dataset. Our method outperforms all previous methods on every frame with much fewer parameters.\\
As explained in Section~\ref{sec:datasets}, some different methods have taken different test sampling strategies. Following \cite{mao2020history}, we choose to test with 256 samples on 22 joints. To make a fair comparison, we evaluate all the methods using the same testing protocol. Our method outperforms all previous methods on every frame with a much less number of parameters.
Besides, previous works usually report short-term ($0\sim500 ms$) and long-term ($500\sim1000 ms$) predictions separately, and \cite{ma2022progressively} reports short-/long- term results using two different models. In our tables, all the results from $0\sim1000 ms$ are predicted by a single model, and for \cite{ma2022progressively}, we report the results of their model which achieves the best performance on long-term prediction. In addition, we also evaluate the two simple approaches mentioned in Section~\ref{intro} on the Human3.6M dataset in Table~\ref{table:tab_res_h36m}: `Repeating Last-frame' takes the last input pose and repeats it $N$ times to serve as output, and `One FC' uses only one single fully connected layer trained on Human3.6m. These results show that the task of human motion prediction could be potentially modeled in a completely different and simple way without explicitly fusing spatial and temporal information. Furthermore, similar to all the previous works, we also detail the action-wise results in Table~\ref{table:tab_res_h36m_act_wise}.

\vspace{2mm}
\paragraph{AMASS and 3DPW datasets.}
In Table~\ref{table:tab_res_amass}, we report the performance of the model trained on AMASS and tested on the AMASS-BMLrub and 3DPW datasets, following the evaluation protocol of \cite{mao2020history}. Different from the Human3.6M dataset where the training and testing data are from the same types of actions performed by different actors, the difference between training and testing data under this protocol is much larger, which makes the task more challenging in terms of generalization. As shown in the table, our approach performs consistently better on long-term prediction. Moreover, our model is much lighter. For example, the parameter size of our model is $\sim 4\%$ of Hisrep~\cite{mao2020history}.

While the commonly used evaluation protocol is to consider the predicted error at different time steps, some works~\cite{sofianos2021space,zhong2022spatio} report their result by taking the average error from the first time step to a certain time step.
We report the predicted error at different time steps in all the tables, except in 
Table~\ref{table:tab_res_h36m_avg}, where we report the average error for comparison with 
\cite{sofianos2021space,zhong2022spatio}. Our approach also achieves better performance than these two methods.

\begin{figure}
    \centering
    \includegraphics[width=0.46\textwidth]{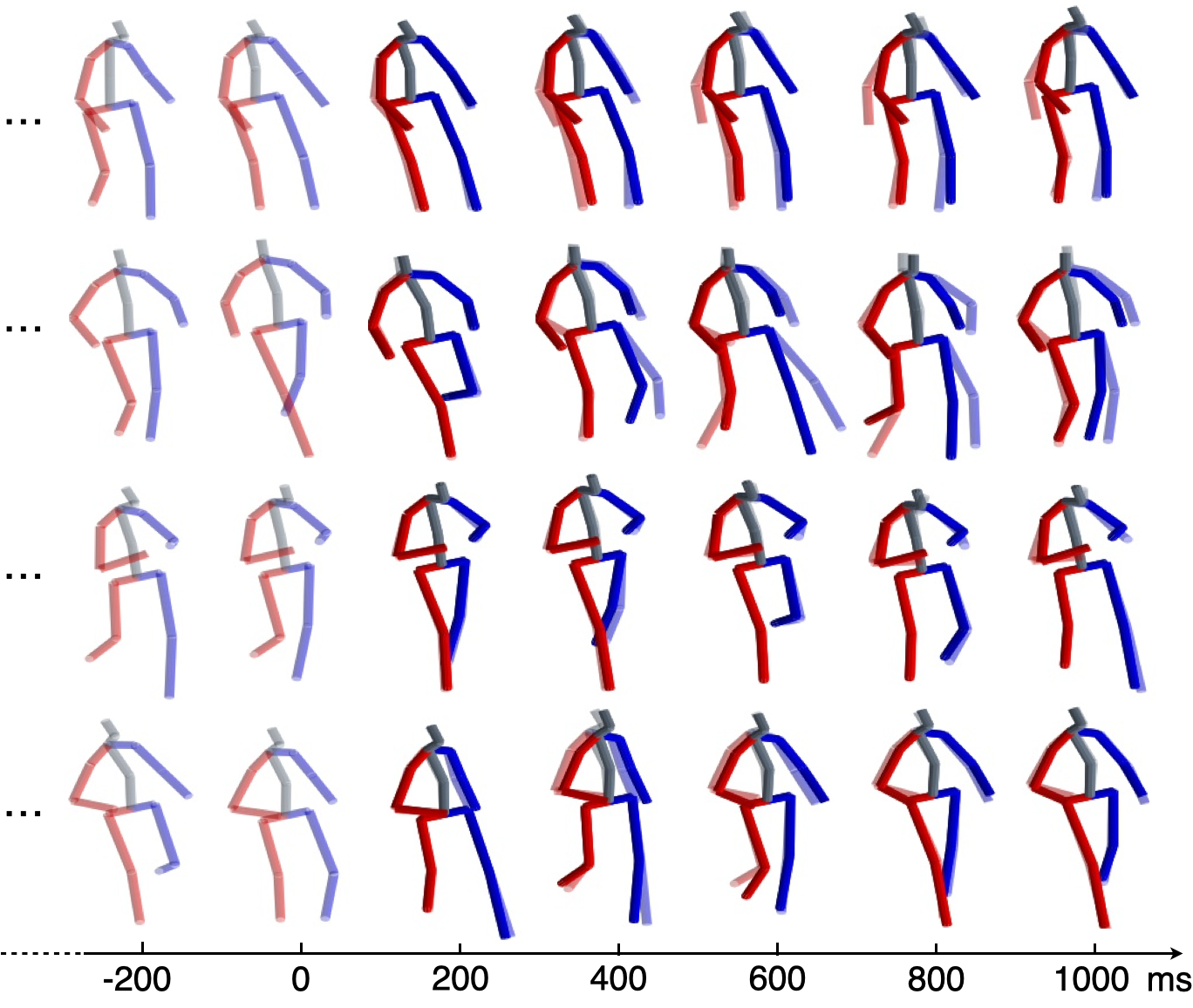}
    \caption{{\bf Qualitative results of our method \method.} The skeletons in light colors are the input (before 0ms) and the ground-truth (after 0ms). Those with dark colors represent the predicted motions. Our prediction results are close to the ground-truth.}
    \label{fig:vis}
    \vspace{-2mm}
\end{figure}

\vspace{2mm}
\paragraph{Qualitative results.} 
In addition to the quantitative results, we provide some qualitative results of our method in Figure~\ref{fig:vis}, showing some testing examples on the Human3.6M dataset. We could find that the predictions of our method perfectly match the ground-truth on short-term prediction, and globally fits the ground-truth on long-term prediction. The error becomes larger when looking into longer predictions, which is a common problem for all the motion prediction methods as shown in Table~\ref{table:tab_res_h36m} and Table~\ref{table:tab_res_amass}. This is because most of the current methods use auto-regression for predicting a longer future, which will make the error accumulate. Moreover, uncertainty grows very quickly with time when predicting human motions.

\subsection{Ablation study}
\label{sec:abl}
We evaluate below the influence of the different components of our approach on the Human3.6M dataset.

\vspace{2mm}
\paragraph{Number of MLP blocks.}
We ablate the of the number of MLP blocks $m$ in Table~\ref{table:tab_abla_layers}. Our proposed architecture already achieves good performance using only 2 MLP blocks with $0.014M$ parameters. The network achieves its best performance with 48 MLP blocks.

\vspace{2mm}
\paragraph{Network architecture.}
In Table~\ref{table:tab_abla_net}, we ablate the different components of our network. 
As the table shows, the temporal feature fusion and layer normalization are both of vital importance to our network. 
If the network just operates along the spatial dimension of the motion sequence without merging any information across different frames, it will lead to degraded results. However, if the network just operates along the temporal dimension, the network will still achieve comparable performance.
Besides, the use of DCT transformation can further improve the performance slightly.

\vspace{2mm}
\paragraph{Data augmentation.}
In Table~\ref{table:tab_abla_aug}, we ablate the use of front-back flip data augmentation and find that the data augmentation slightly improves the performance.

\vspace{2mm}
\paragraph{Loss.}
In Table~\ref{table:tab_abla_losses}, we evaluate the importance of different loss terms used during training. As shown in the table, with the help of the velocity loss $\mathcal{L}_{v}$, the network achieves better performance on long-term predictions while maintaining the same performance on the short-term.

\vspace{2mm}
\paragraph{Learning residual displacement.}
In Table~\ref{table:tab_abla_res}, we analyze the importance of the proposed residual displacement and compare it to other types of residual used in previous works~\cite{martinez2017human,mao2019learning}. Our method aims to predict the differences between each future pose and the last observed pose, after the IDCT transformation. When predicting directly the absolute 3D pose (`w/o residual'), the performance drops dramatically. We also test other types of residual by either learning the residual in the DCT space, before applying the IDCT transformation (`Before IDCT') following~\cite{mao2019learning}, or learning the velocity of the motion (`consecutive') following~\cite{martinez2017human}, and both achieve inferior performance compared to our proposed residual displacement.

\setlength{\tabcolsep}{1.4pt}

\begin{table}[t!]
\vspace{2mm}
\caption{\textbf{Ablation of different loss terms} on Human3.6M.}\vspace{2mm}
\label{table:tab_abla_losses}
\centering
\resizebox{0.48\textwidth}{!}{\setlength{\tabcolsep}{1.6mm}{
\begin{tabular}{cc|cccccccc}

\toprule
$\mathcal{L}_{re}$ & $\mathcal{L}_v$ & 80 & 160 & 320 & 400 & 560 & 720 & 880 & 1000 \\
\midrule
\checkmark &  &  9.6 & 21.8 & 46.5 & 57.5 & 76.7 & 91.5 & 103.5 & 111.3\\
\rowcolor{gray!15} \cellcolor{white}\checkmark & \cellcolor{white} \checkmark & \bf 9.6 & \bf 21.7 & \bf 46.3 & \bf 57.3 & \bf 75.7 & \bf 90.1 & \bf 101.8 & \bf 109.4\\

\bottomrule
\end{tabular}}}

\end{table}
\setlength{\tabcolsep}{1.4pt}
\begin{table}[t!]
\caption{\textbf{Analysis of different types of residual displacement} on Human3.6M.
\textit{\method} predicts the differences of each future frame with the last observation (after IDCT). 
\textit{'Before IDCT'} learns the residual before applying the IDCT transformation. 
\textit{'Consecutive'} learns the velocity between consecutive frames. 
\textit{'w/o residual'} predicts directly the absolute 3D poses.
}
\vspace{2mm}
\label{table:tab_abla_res}
\centering
\rowcolors{3}{gray!15}{white}
\resizebox{0.48\textwidth}{!}{\setlength{\tabcolsep}{0.8mm}{
\begin{tabular}{l|cccccccc}
\toprule
Residual & 80 & 160 & 320 & 400 & 560 & 720 & 880 & 1000 \\
\midrule
\cellcolor{white}w/o residual & 12.4 & 25.1 & 50.7 & 61.6 & 80.1 & 93.9 & 105.5 & 113.0 \\
\cellcolor{white}Consecutive & 9.7  & 22.0  & 46.8  & 57.8 &  76.5 &  90.7  & 102.4  & 110.1 \\
\cellcolor{white}Before IDCT & 10.4  & 23.0  & 48.2 &  59.1 &  77.9 &  91.8  & 103.2  & 110.5 \\
\midrule
\cellcolor{white}\method(ours) & \bf 9.6 & \bf 21.7 & \bf 46.3 & \bf 57.3 & \bf 75.7 & \bf 90.1 & \bf 101.8 & \bf 109.4\\

\bottomrule
\end{tabular}}}
\vspace{-5mm}
\end{table}

\section{Conclusion}
In this paper, we present \method, a simple-yet-effective network for human motion prediction. \method is composed of only fully connected layers, layer normalization, and transpose operations. The only non-linear operation is thus the layer normalization. While using much fewer parameters, \method achieves state-of-the-art performance on various benchmarks. The reported ablation study also demonstrates the interest of various design choices, highlighting the importance of temporal information fusion in this task. We hope the simplicity of \method will help the community to rethink the task of human motion prediction.

\section{Acknowledgement}
This research was supported by ANR-3IA MIAI (ANR-19-P3IA-0003), ANR-JCJC ML3RI (ANR-19-CE33-0008-01), H2020 SPRING (funded by EC under GA \#871245), by the Spanish government with the project MoHuCo PID2020-120049RB-I00 and by an Amazon Research Award. This project has received funding from the CHIST-ERA IPALM project.

{\small
\bibliographystyle{ieee_fullname}
\bibliography{egbib}

\begin{thebibliography}{10}\itemsep=-1pt

\bibitem{PosePrior_Akhter:CVPR:2015}
Ijaz Akhter and Michael~J. Black.
\newblock {Pose-Conditioned Joint Angle Limits for {3D} Human Pose
  Reconstruction}.
\newblock In {\em Conference on Computer Vision and Pattern Recognition}, 2015.

\bibitem{aksan2021spatio}
Emre Aksan, Manuel Kaufmann, Peng Cao, and Otmar Hilliges.
\newblock {A Spatio-Temporal Transformer for 3D Human Motion Prediction}.
\newblock In {\em 3DV}, 2021.

\bibitem{DanceDB:Aristidou:2019}
Andreas Aristidou, Ariel Shamir, and Yiorgos Chrysanthou.
\newblock {Digital Dance Ethnography: {O}rganizing Large Dance Collections}.
\newblock {\em J. Comput. Cult. Herit.}, 12(4), Nov. 2019.

\bibitem{ba2016layer}
Jimmy~Lei Ba, Jamie~Ryan Kiros, and Geoffrey~E. Hinton.
\newblock {Layer Normalization}.
\newblock In {\em arXiv Preprint}, 2016.

\bibitem{dfaust:CVPR:2017}
Federica Bogo, Javier Romero, Gerard Pons-Moll, and Michael~J. Black.
\newblock {Dynamic {FAUST}: Registering Human Bodies in Motion}.
\newblock In {\em Conference on Computer Vision and Pattern Recognition}, 2017.

\bibitem{bouazizi2022motionmixer}
Arij Bouazizi, Adrian Holzbock, Ulrich Kressel, Klaus Dietmayer, and Vasileios
  Belagiannis.
\newblock Motionmixer: Mlp-based 3d human body pose forecasting.
\newblock {\em arXiv preprint arXiv:2207.00499}, 2022.

\bibitem{brand2000style}
Matthew Brand and Aaron Hertzmann.
\newblock {Style Machines}.
\newblock In {\em Computer Graphics and Interactive Techniques}, 2000.

\bibitem{butepage2017deep}
Judith Butepage, Michael~J. Black, Danica Kragic, and Hedvig Kjellstrom.
\newblock {Deep Representation Learning for Human Motion Prediction and
  Classification}.
\newblock In {\em Conference on Computer Vision and Pattern Recognition}, 2017.

\bibitem{butepage2018anticipating}
Judith B\"utepage, Hedvig Kjellstr\"om, and Danica Kragic.
\newblock {Anticipating Many Futures: Online Human Motion Prediction and
  Generation for Human-Robot Interaction}.
\newblock In {\em International Conference on Robotics and Automation}, 2018.

\bibitem{cai2020learning}
Yujun Cai, Lin Huang, Yiwei Wang, Tat-Jen Cham, Jianfei Cai, Junsong Yuan, Jun
  Liu, Xu Yang, Yiheng Zhu, Xiaohui Shen, and Others.
\newblock {Learning Progressive Joint Propagation for Human Motion Prediction}.
\newblock In {\em European Conference on Computer Vision}, 2020.

\bibitem{chatzitofis2020human4d}
Anargyros Chatzitofis, Leonidas Saroglou, Prodromos Boutis, Petros Drakoulis,
  Nikolaos Zioulis, Shishir Subramanyam, Bart Kevelham, Caecilia Charbonnier,
  Pablo Cesar, Dimitrios Zarpalas, and Others.
\newblock {HUMAN4D: A Human-Centric Multimodal Dataset for Motions and
  Immersive Media}.
\newblock In {\em American Control Conference}, 2020.

\bibitem{chiu2019action}
Hsu-kuang Chiu, Ehsan Adeli, Borui Wang, De-An Huang, and Juan~Carlos Niebles.
\newblock {Action-Agnostic Human Pose Forecasting}.
\newblock In {\em IEEE Winter Conference on Applications of Computer Vision},
  2019.

\bibitem{dang2021msr}
Lingwei Dang, Yongwei Nie, Chengjiang Long, Qing Zhang, and Guiqing Li.
\newblock {MSR-GCN: Multi-Scale Residual Graph Convolution Networks for Human
  Motion Prediction}.
\newblock In {\em International Conference on Computer Vision}, 2021.

\bibitem{ACCAD}
Advanced Computing~Center for~the Arts and Design.
\newblock {ACCAD MoCap Dataset}.

\bibitem{fragkiadaki2015recurrent}
Katerina Fragkiadaki, Sergey Levine, Panna Felsen, and Jitendra Malik.
\newblock {Recurrent Network Models for Human Dynamics}.
\newblock In {\em International Conference on Computer Vision}, 2015.

\bibitem{ghorbani2020movi}
Saeed Ghorbani, Kimia Mahdaviani, Anne Thaler, Konrad Kording, Douglas~James
  Cook, Gunnar Blohm, and Nikolaus~F. Troje.
\newblock {{MoVi}: A Large Multipurpose Motion and Video Dataset}, 2020.

\bibitem{gong2011multi}
Haifeng Gong, Jack Sim, Maxim Likhachev, and Jianbo Shi.
\newblock {Multi-Hypothesis Motion Planning for Visual Object Tracking}.
\newblock In {\em International Conference on Computer Vision}, 2011.

\bibitem{gui2018adversarial}
Liang-Yan Gui, Yu-Xiong Wang, Xiaodan Liang, and Jos\'e~MF Moura.
\newblock {Adversarial Geometry-Aware Human Motion Prediction}.
\newblock In {\em European Conference on Computer Vision}, 2018.

\bibitem{guomulti}
Wen Guo, Xiaoyu Bie, Xavier Alameda-Pineda, and Francesc Moreno-Noguer.
\newblock {Multi-Person Extreme Motion Prediction}.
\newblock In {\em Conference on Computer Vision and Pattern Recognition}, 2022.

\bibitem{hernandeziccv2019}
Alejandro Hernandez, Jurgen Gall, and Francesc Moreno-Noguer.
\newblock {Human Motion Prediction via Spatio-Temporal Inpainting}.
\newblock In {\em International Conference on Computer Vision}, 2019.

\bibitem{TCD_hands}
Ludovic Hoyet, Kenneth Ryall, Rachel Mcdonnell, and Carol O'sullivan.
\newblock {Sleight of Hand: Perception of Finger Motion from Reduced Marker
  Sets}.
\newblock In {\em ACM SIGGRAPH}, 2012.

\bibitem{hu2018squeeze}
Jie Hu, Li Shen, and Gang Sun.
\newblock Squeeze-and-excitation networks.
\newblock In {\em Proceedings of the IEEE conference on computer vision and
  pattern recognition}, pages 7132--7141, 2018.

\bibitem{ionescu2013human3}
Catalin Ionescu, Dragos Papava, Vlad Olaru, and Cristian Sminchisescu.
\newblock {Human3. 6m: Large Scale Datasets and Predictive Methods for 3D Human
  Sensing in Natural Environments}.
\newblock {\em IEEE Transactions on Pattern Analysis and Machine Intelligence},
  2013.

\bibitem{jain2016structural}
Ashesh Jain, Amir~R. Zamir, Silvio Savarese, and Ashutosh Saxena.
\newblock {Structural-Rnn: Deep Learning on Spatio-Temporal Graphs}.
\newblock In {\em Conference on Computer Vision and Pattern Recognition}, 2016.

\bibitem{kingma2014adam}
Diederik~P. Kingma and Jimmy Ba.
\newblock {Adam: A Method for Stochastic Optimization}.
\newblock In {\em arXiv Preprint}, 2014.

\bibitem{kipf2016semi}
Thomas~N. Kipf and Max Welling.
\newblock {Semi-Supervised Classification with Graph Convolutional Networks}.
\newblock In {\em International Conference for Learning Representations}, 2017.

\bibitem{koppula2013anticipating}
Hema~Swetha Koppula and Ashutosh Saxena.
\newblock {Anticipating Human Activities for Reactive Robotic Response}.
\newblock In {\em International Conference on Intelligent Robots and Systems},
  2013.

\bibitem{BMLhandball}
Bio~Motion Lab.
\newblock {BMLhandball Motion Capture Database}.

\bibitem{lebailly2020motion}
Tim Lebailly, Sena Kiciroglu, Mathieu Salzmann, Pascal Fua, and Wei Wang.
\newblock {Motion Prediction Using Temporal Inception Module}.
\newblock In {\em Asian Conference on Computer Vision}, 2020.

\bibitem{lehrmann2014efficient}
Andreas~M. Lehrmann, Peter~V. Gehler, and Sebastian Nowozin.
\newblock {Efficient Nonlinear Markov Models for Human Motion}.
\newblock In {\em Conference on Computer Vision and Pattern Recognition}, 2014.

\bibitem{li2018convolutional}
Chen Li, Zhen Zhang, Wee~Sun Lee, and Gim~Hee Lee.
\newblock {Convolutional Sequence to Sequence Model for Human Dynamics}.
\newblock In {\em Conference on Computer Vision and Pattern Recognition}, 2018.

\bibitem{li2021symbiotic}
Maosen Li, Siheng Chen, Xu Chen, Ya Zhang, Yanfeng Wang, and Qi Tian.
\newblock {Symbiotic Graph Neural Networks for 3D Skeleton-Based Human Action
  Recognition and Motion Prediction}.
\newblock {\em IEEE Transactions on Pattern Analysis and Machine Intelligence},
  2021.

\bibitem{li2020dynamic}
Maosen Li, Siheng Chen, Yangheng Zhao, Ya Zhang, Yanfeng Wang, and Qi Tian.
\newblock {Dynamic Multiscale Graph Neural Networks for 3D Skeleton Based Human
  Motion Prediction}.
\newblock In {\em Conference on Computer Vision and Pattern Recognition}, 2020.

\bibitem{liu2019towards}
Zhenguang Liu, Shuang Wu, Shuyuan Jin, Qi Liu, Shijian Lu, Roger Zimmermann,
  and Li Cheng.
\newblock {Towards Natural and Accurate Future Motion Prediction of Humans and
  Animals}.
\newblock In {\em Conference on Computer Vision and Pattern Recognition}, 2019.

\bibitem{MoSh_lopermahmoodetal2014}
Matthew Loper, Naureen Mahmood, and Michael~J. Black.
\newblock {{MoSh}: {Motion} And {Shape Capture} from {Sparse Markers}}.
\newblock {\em IEEE Transactions on Robotics and Automation}, 33(6), Nov. 2014.

\bibitem{loper2015smpl}
Matthew Loper, Naureen Mahmood, Javier Romero, Gerard Pons-Moll, and Michael~J.
  Black.
\newblock {SMPL: A Skinned Multi-Person Linear Model}.
\newblock {\em IEEE Transactions on Robotics and Automation}, 34(6), 2015.

\bibitem{Eyes_Japan}
Eyes JAPAN~Co Ltd.
\newblock {Eyes Japan MoCap Dataset}.

\bibitem{ma2022progressively}
Tiezheng Ma, Yongwei Nie, Chengjiang Long, Qing Zhang, and Guiqing Li.
\newblock {Progressively Generating Better Initial Guesses Towards Next Stages
  for High-Quality Human Motion Prediction}.
\newblock In {\em Conference on Computer Vision and Pattern Recognition}, 2022.

\bibitem{mahmood2019amass}
Naureen Mahmood, Nima Ghorbani, Nikolaus~F. Troje, Gerard Pons-Moll, and
  Michael~J. Black.
\newblock {{AMASS}: Archive of Motion Capture as Surface Shapes}.
\newblock In {\em International Conference on Computer Vision}, 2019.

\bibitem{mandery2015kit}
Christian Mandery, {\"O}mer Terlemez, Martin Do, Nikolaus Vahrenkamp, and Tamim
  Asfour.
\newblock {The KIT Whole-Body Human Motion Database}.
\newblock In {\em ICAR}, 2015.

\bibitem{mao2020history}
Wei Mao, Miaomiao Liu, and Mathieu Salzmann.
\newblock {History Repeats Itself: Human Motion Prediction via Motion
  Attention}.
\newblock In {\em European Conference on Computer Vision}, 2020.

\bibitem{mao2019learning}
Wei Mao, Miaomiao Liu, Mathieu Salzmann, and Hongdong Li.
\newblock {Learning Trajectory Dependencies for Human Motion Prediction}.
\newblock In {\em International Conference on Computer Vision}, 2019.

\bibitem{martinez2017human}
Julieta Martinez, Michael~J. Black, and Javier Romero.
\newblock {On Human Motion Prediction Using Recurrent Neural Networks}.
\newblock In {\em Conference on Computer Vision and Pattern Recognition}, 2017.

\bibitem{MPI_HDM05}
M. M\"{u}ller, T. R\"{o}der, M. Clausen, B. Eberhardt, B. Kr\"{u}ger, and A.
  Weber.
\newblock {Documentation Mocap Database {HDM05}}.
\newblock Technical Report CG-2007-2, Universit\"{a}t Bonn, June 2007.

\bibitem{nair2010rectified}
Vinod Nair and Geoffrey~E. Hinton.
\newblock {Rectified Linear Units Improve Restricted Boltzmann Machines}.
\newblock In {\em International Conference on Machine Learning}, 2010.

\bibitem{paden2016survey}
Brian Paden, Michal {\v{C}}\'ap, Sze~Zheng Yong, Dmitry Yershov, and Emilio
  Frazzoli.
\newblock {A Survey of Motion Planning and Control Techniques for Self-Driving
  Urban Vehicles}.
\newblock {\em IEEE Transactions on Robotics and Automation}, 2016.

\bibitem{paszke2019pytorch}
Adam Paszke, Sam Gross, Francisco Massa, Adam Lerer, James Bradbury, Gregory
  Chanan, Trevor Killeen, Zeming Lin, Natalia Gimelshein, Luca Antiga, and
  Others.
\newblock {Pytorch: An Imperative Style, High-Performance Deep Learning
  Library}.
\newblock In {\em Advances in Neural Information Processing Systems}, 2019.

\bibitem{sigal2010humaneva}
Leonid Sigal, Alexandru~O. Balan, and Michael~J. Black.
\newblock {{HumanEva}: Synchronized Video and Motion Capture Dataset and
  Baseline Algorithm for Evaluation of Articulated Human Motion}.
\newblock {\em International Journal of Computer Vision}, 2010.

\bibitem{sofianos2021space}
Theodoros Sofianos, Alessio Sampieri, Luca Franco, and Fabio Galasso.
\newblock {Space-Time-Separable Graph Convolutional Network for Pose
  Forecasting}.
\newblock In {\em International Conference on Computer Vision}, 2021.

\bibitem{sperduti1997supervised}
Alessandro Sperduti and Antonina Starita.
\newblock {Supervised Neural Networks for the Classification of Structures}.
\newblock {\em IEEE Transactions on Robotics and Automation}, 1997.

\bibitem{taylor2007modeling}
Graham~W. Taylor, Geoffrey~E. Hinton, and Sam~T. Roweis.
\newblock {Modeling Human Motion Using Binary Latent Variables}.
\newblock In {\em Advances in Neural Information Processing Systems}, 2007.

\bibitem{tolstikhin2021mlp}
Ilya~O Tolstikhin, Neil Houlsby, Alexander Kolesnikov, Lucas Beyer, Xiaohua
  Zhai, Thomas Unterthiner, Jessica Yung, Andreas Steiner, Daniel Keysers,
  Jakob Uszkoreit, et~al.
\newblock Mlp-mixer: An all-mlp architecture for vision.
\newblock {\em Advances in Neural Information Processing Systems},
  34:24261--24272, 2021.

\bibitem{BMLrub}
Nikolaus~F. Troje.
\newblock {Decomposing Biological Motion: {A} Framework for Analysis and
  Synthesis of Human Gait Patterns}.
\newblock {\em Journal of Vision}, 2(5), Sept. 2002.

\bibitem{TotalCapture_Trumble}
Matt Trumble, Andrew Gilbert, Charles Malleson, Adrian Hilton, and John
  Collomosse.
\newblock {Total Capture: 3D Human Pose Estimation Fusing Video and Inertial
  Sensors}.
\newblock In {\em British Machine Vision Conference}, 2017.

\bibitem{cmuWEB}
Carnegie~Mellon University.
\newblock {CMU MoCap Dataset}.

\bibitem{SFU}
Simon~Fraser University and National University~of Singapore.
\newblock {SFU Motion Capture Database}.

\bibitem{vaswani2017attention}
Ashish Vaswani, Noam Shazeer, Niki Parmar, Jakob Uszkoreit, Llion Jones,
  Aidan~N. Gomez, Lukasz Kaiser, and Illia Polosukhin.
\newblock {Attention Is All You Need}.
\newblock In {\em Advances in Neural Information Processing Systems}, 2017.

\bibitem{vonMarcard2018}
Timo Von~Marcard, Roberto Henschel, Michael~J. Black, Bodo Rosenhahn, and
  Gerard Pons-Moll.
\newblock {Recovering Accurate 3D Human Pose in the Wild Using IMUs and a
  Moving Camera}.
\newblock In {\em European Conference on Computer Vision}, 2018.

\bibitem{wang2005gaussian}
Jack~M. Wang, David~J. Fleet, and Aaron Hertzmann.
\newblock {Gaussian Process Dynamical Models}.
\newblock In {\em Advances in Neural Information Processing Systems}, 2005.

\bibitem{wang2007gaussian}
Jack~M. Wang, David~J. Fleet, and Aaron Hertzmann.
\newblock {Gaussian Process Dynamical Models for Human Motion}.
\newblock {\em IEEE Transactions on Pattern Analysis and Machine Intelligence},
  2007.

\bibitem{zhong2022spatio}
Chongyang Zhong, Lei Hu, Zihao Zhang, Yongjing Ye, and Shihong Xia.
\newblock {Spatio-Temporal Gating-Adjacency GCN For Human Motion Prediction}.
\newblock In {\em Conference on Computer Vision and Pattern Recognition}, 2022.

\end{thebibliography}
}


\end{document}